\documentclass[runningheads]{llncs}
\usepackage[T1]{fontenc}
\usepackage{graphicx}
\usepackage{subcaption}
\usepackage{amsmath} 
\usepackage{amsfonts}
\usepackage{booktabs}
\usepackage{multirow}
\usepackage{makecell}
\usepackage{float}
\usepackage{makecell}

\usepackage{xcolor}
\definecolor{green}{RGB}{0,128,0}

\graphicspath{ {./figures/} }

\begin{document}
\title{OASIC: Occlusion-Agnostic and Severity-Informed Classification}
\author{Kay Gijzen\inst{1,2} \and
Gertjan J. Burghouts\inst{2} \and
Daniël M. Pelt\inst{1}}
\authorrunning{K. Gijzen, et al.}
\institute{Leiden University \and TNO}
\maketitle % typeset the header of the contribution
\begin{abstract}

Severe occlusions of objects pose a major challenge for computer vision. We show that two root causes are (1) the loss of visible information and (2) the distracting patterns caused by the occluders. Our approach addresses both causes at the same time. First, the distracting patterns are removed at test-time, via masking of the occluding patterns. This masking is independent of the type of occlusion, by handling the occlusion through the lens of visual anomalies w.r.t. the object of interest. Second, to deal with less visual details, we follow standard practice by masking random parts of the object during training~\cite{kumar2017hide}, for various degrees of occlusions. We discover that (a) it is possible to estimate the degree of the occlusion (i.e. severity) at test-time, and (b) that a model optimized for a specific degree of occlusion also performs best on a similar degree during test-time. Combining these two insights brings us to a severity-informed classification model called OASIC: Occlusion Agnostic Severity Informed Classification. We estimate the severity of occlusion for a test image, mask the occluder, and select the model that is optimized for the degree of occlusion. This strategy performs better than any single model optimized for any smaller or broader range of occlusion severities. Experiments show that combining gray masking with adaptive model selection improves $\text{AUC}_\text{occ}$ by +18.5 over standard training on occluded images and +23.7 over finetuning on unoccluded images. 

\end{abstract}

\section{Introduction}
% hook → context → problem → research questions → contributions → structure
% \textbf{Hook.} +
Modern computer vision models perform impressively on clean, fully visible images. In practice, however, objects are often partially hidden or obscured, e.g., foliage and smoke. Finegrained classification is challenging under severe occlusions, because less of the details are visible which are needed to distinguish the classes. We consider up to 90\% occlusion, i.e., 10\% of the finegrained class is visible. Finegrained classification under such severe occlusions is a under-explored area, but essential for real-world deployment. 

The gap between clean training images and occluded real-world conditions is a significant challenge in computer vision~\cite{shen2025performance,torralba2011unbiased}. Occlusion is challenging for two reasons: it reduces the visible regions of an object and introduces distracting patterns that can mislead models \cite{zhang2018deepvoting,shorten2019survey}. This is confirmed for finegrained classification, by the findings in this paper. Datasets containing annotated occlusion are scarce, and real-world occlusions are often unpredictable. That makes robust finegrained classification under severe occlusions a particularly difficult yet interesting problem.

To address the challenges posed by occlusion, we introduce OASIC (Occlusion-Agnostic Severity-Informed Classification), a method designed to explicitly tackle the two fundamental difficulties of occlusion: (i) the loss of visible information and (ii) the visual distraction introduced by the occluders themselves. OASIC handles occlusion as a form of visual anomaly with respect to the object of interest, allowing it to handle any type of occlusion—such as vegetation, smoke, or other visual obstructions—without requiring prior knowledge of the occluder. Our method employs pixel-level occlusion likelihood maps, enabling what we term occlusion-agnostic segmentation. From these maps, we derive segmentation masks that serve two complementary roles. First, they mitigate visual distraction by replacing occluded regions with a neutral tone, effectively masking out the misleading texture cues introduced by the occluders. Second, the same segmentation maps allow us to estimate the severity of occlusion, defined as the fraction of the image affected by occlusion. During training, OASIC follows standard practice by applying random masks to simulate varying degrees of occlusion, ensuring robustness across different severity levels. Through this process, we discover two key properties: (a) the degree of occlusion can be estimated reliably at test time, and (b) a model optimized for a specific severity level performs best when tested on similar levels of occlusion. Combining these insights, OASIC operates as a severity-informed classification framework. At inference time, the method estimates the occlusion severity for each test image, masks the occluding patterns, and dynamically selects the model trained for the corresponding degree of occlusion. This strategy consistently outperforms any single model trained on a fixed or broader range of occlusion severities, achieving robust and adaptable performance under diverse occlusion conditions.

The main contributions of this work are threefold. (1) We introduce OASIC, a unified framework that addresses both the loss of visual information and the distracting patterns caused by occluders. (2) Occlusion-agnostic segmentation by interpreting occluded regions as visual anomalies, enabling the localization of arbitrary occluders without prior knowledge of their appearance, suppressing distraction. (3) A severity-informed model selection strategy that leverages the estimated occlusion severity to dynamically select the most suitable model from a set of models which were trained across varying occlusion levels. This adaptive approach leads to consistently improved classification robustness under diverse and unpredictable occlusion conditions.

\section{Related Work}\label{chap:related}

To handle occlusion, data augmentation techniques create modified training samples that discourage reliance on specific image regions. Mixup~\cite{zhang2017mixup} blends two images through pixel averaging, while CutMix~\cite{yun2019cutmix} replaces patches from one image with another. Hide-and-Seek~\cite{kumar2017hide} randomly hides image regions to force attention to alternative cues. These methods promote part-based learning by exposing models to partial views. TransMix~\cite{chen2022transmix}, extending CutMix, uses transformer attention to weight label contributions, further enhancing part-based robustness. Such augmentations improve recognition under occlusion by linking labels to visible regions, but they mainly mitigate information loss and do not fully resolve confusion from misleading occluder textures.

Part-based methods explicitly model object structure. CompositionalNets~\cite{kortylewski2020compositional} include an occlusion localization module and part dictionaries that reason over visibility, improving recognition of partially visible objects. TDMPNet~\cite{xiao2020tdmpnet} estimates visibility maps and suppresses occluded features via top-down attention, yielding cleaner feature representations. Despite their strengths, these CNN-based methods remain limited by local receptive fields and poor long-range reasoning, making them sensitive to structured occlusion~\cite{fawzi2016measuring,devries2017improved}.

Transformer-based models (ViTs) demonstrate superior robustness to occlusion~\cite{naseer2021intriguing,kassaw2025deep}, largely due to self-supervised pretraining. Masked Image Modeling (MIM)~\cite{kong2023understanding}, as used in MAE~\cite{he2022masked} and iBOT~\cite{zhou2021ibot}, trains models to reconstruct masked regions, inherently promoting occlusion resilience. Large-scale vision transformers such as CLIP~\cite{radford2021learning} and DINOv2~\cite{oquab2024dinov2} further benefit from massive, diverse pretraining, yielding general-purpose visual representations adaptable to many downstream tasks. However, transformer robustness is largely incidental: while ViTs retain accuracy under partial occlusion~\cite{naseer2021intriguing}, their performance degrades under strong or structured occlusions~\cite{kassaw2025deep}. This highlights the need for complementary strategies beyond pretraining to achieve reliable occlusion handling.

To our knowledge, no prior work addresses occlusion segmentation in an \textit{occlusion-agnostic} manner using pixel-level anomaly maps. We repurpose anomaly detection methods such as \textit{AnomalyDINO}~\cite{damm2025anomalydino}, PatchCore~\cite{roth2022towards}, and DRAEM~\cite{zavrtanik2021draem}, which localize visual irregularities via per-pixel anomaly likelihoods. By interpreting high anomaly scores as occluded regions, we achieve occlusion segmentation without explicit supervision, relying only on image-level labels. Unlike segmentation models such as SAM2~\cite{ravi2024sam} or OVSeg~\cite{liang2023open}, which require prior knowledge or prompting for each occluder type~\cite{vashisht2025effective}, our approach generalizes across arbitrary occlusions through an occlusion-agnostic formulation.

\section{OASIC}\label{chap:method}

We present our method, Occlusion-Agnostic Severity-Informed Classification (OASIC). We start with a high-level overview of the approach, followed by detailed explanations of its main components: occlusion map generation, occlusion segmentation, severity estimation, and severity-informed model selection.

First, occluded regions are localized using visual anomaly detection and subsequently masked with a neutral gray tone, replacing distracting textures with a uniform appearance. This reduces visual noise and allows the model to focus on visible, relevant object features. Second, from the occlusion likelihoods in the image, we estimate occlusion severity — the fraction of the image affected by occlusion — which guides the selection of the most suitable model. That model is selected from a pool of models which were finetuned for various levels of occlusions. This adaptive strategy maintains stable classification performance across varying visibility conditions. Our experiments confirm that no single model performs optimally across all occlusion severities, which emphasizes the need for severity-informed model selection.

\begin{figure}[h]
    % \vspace{7.5mm}
    \centering
    \includegraphics[width=\textwidth]{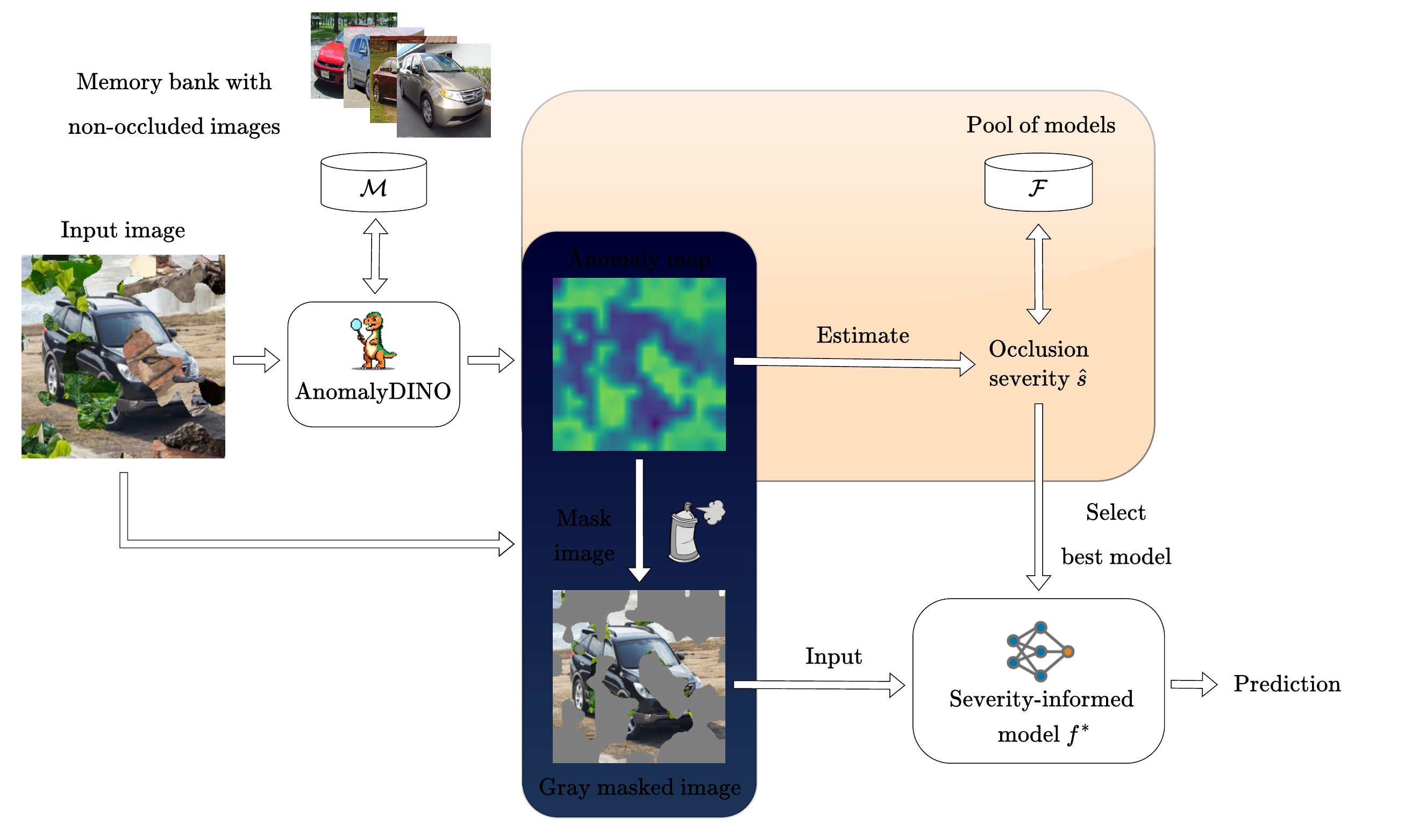}
        \caption{\textbf{Occlusion handling by OASIC.} 
At test time, an occlusion map is inferred by scoring against the memory bank $\mathcal{M}$. The segmented mask is turned into gray to suppress distraction, while the estimated occlusion severity informs the selection of the most suitable classification model $f^\ast$ from the pool $\mathcal{F}$, to better handle reduced visual information.}
    \label{fig:framework}
\end{figure}

\subsection{Occlusion segmentation and masking}

OASIC treats occlusion as a visual anomaly relative to the object of interest, enabling the localization of diverse occluders—such as vegetation, smoke, or other visual obstructions—without requiring prior knowledge of their appearance. We employ \textit{AnomalyDINO}~\cite{damm2025anomalydino} to obtain occlusion likelihoods. As an occlusion-agnostic method, it detects irregularities in appearance rather than predefined occluder types, producing anomaly maps that we interpret as per-pixel occlusion probabilities.

Based on clean reference images without occlusion, AnomalyDINO produces an anomaly map $A \in [0,1]^{H \times W}$ that assigns each pixel a likelihood of being occluded. To obtain a discrete representation of these regions, the map is thresholded at a value $\tau$, yielding a binary occlusion map $O \in \{0,1\}^{H \times W}$ that indicates which pixels are classified as occluded:

\begin{equation}
O_{i,j} =
\begin{cases}
1, & \text{if } A_{i,j} \geq \tau, \\
0, & \text{otherwise},
\end{cases}
\label{eq:threshold}
\end{equation}

where $\tau \in [0,1]$. The choice of threshold $\tau$ allows control over the aggressiveness of occlusion detection: \textit{aggressive detection} uses a low threshold to capture as much occlusion as possible at the cost of potential false positives, while \textit{conservative detection} uses a high threshold to mark only high-confidence occluded pixels, reducing false positives.

\subsubsection{Adaptive thresholding} 
Instead of a fixed threshold~$\tau$, we adopt Otsu's method~\cite{otsu1979threshold}, which analyzes the histogram of anomaly scores in~$A$ and selects the threshold~$\tau^*$ that maximizes the between-class variance. This adapts the thresholded binary occlusion map~$O$ to each image.  
Let $p(i)$ be the normalized histogram of anomaly values for intensity levels $i \in \{0,\dots,L-1\}$. The class probabilities and means for a threshold~$t$ are:

\begin{align}
\omega_0(t) &= \sum_{i=0}^{t} p(i), &
\omega_1(t) &= \sum_{i=t+1}^{L-1} p(i), \\
\mu_0(t) &= \frac{1}{\omega_0(t)} \sum_{i=0}^{t} i\,p(i), &
\mu_1(t) &= \frac{1}{\omega_1(t)} \sum_{i=t+1}^{L-1} i\,p(i).
\end{align}
\\
The between-class variance is then given by:
\begin{equation}
\sigma_b^2(t) = \omega_0(t)\,\omega_1(t)\,[\mu_0(t) - \mu_1(t)]^2.
\end{equation}
\\
Otsu’s method selects the optimal threshold as:
\begin{equation}
\tau^* = \arg\max_t \sigma_b^2(t).
\end{equation}
\\
Finally, the binary occlusion map~$O$ is obtained through Equation~\ref{eq:threshold}, using $\tau^*$ as the threshold $\tau$.
% \begin{equation}
% O_{i,j} =
% \begin{cases}
% 1, & \text{if } A_{i,j} \geq \tau^*,\\[4pt]
% 0, & \text{otherwise.}
% \end{cases}
% \end{equation}

\subsubsection{Occlusion masking} 
\begin{figure}[h]
    \centering

    \begin{subfigure}[t]{0.22\textwidth}
        \includegraphics[width=\linewidth]{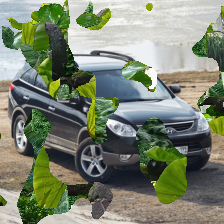}
        \caption{Occluded image}
        \label{fig:masking_occluded}
    \end{subfigure}
    \hfill
    \begin{subfigure}[t]{0.22\textwidth}
        \includegraphics[width=\linewidth]{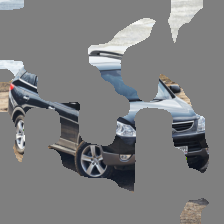}
        \caption{Masked image, $\tau = 0.3$}
        \label{fig:mask03}
    \end{subfigure}
    \hfill
    \begin{subfigure}[t]{0.22\textwidth}
        \includegraphics[width=\linewidth]{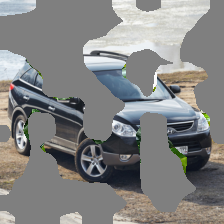}
        \caption{Masked image, $\tau = 0.5$}
        \label{fig:mask05}
    \end{subfigure}
    \hfill
    \begin{subfigure}[t]{0.22\textwidth}
        \includegraphics[width=\linewidth]{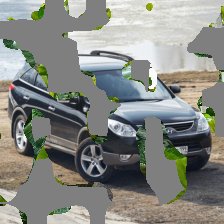}
        \caption{Masked image, $\tau = 0.7$}
        \label{fig:mask07}
    \end{subfigure}

    \caption{\textbf{Comparison of an occluded image and its masked versions at different thresholds $\tau$.} From left to right: the original occluded image, and masks applied with thresholds 0.3, 0.5, and 0.7.}
    \label{fig:mask_comparison}
\end{figure}

Using the binary occlusion map $O$, we mask occluded regions with a uniform gray value. This suppresses the high-frequency textures typically caused by occlusions, preventing them from interfering with feature extraction and classification. Specifically, we construct a masked image $I_{\text{masked}}$ by replacing all pixels identified as occluded ($O_{i,j} = 1$) with a uniform gray value $g$:

\begin{equation}
I_{\text{masked},i,j} =
\begin{cases}
g, & \text{if } O_{i,j} = 1, \\
I_{i,j}, & \text{otherwise},
\end{cases}
\end{equation}

where $I$ denotes the original image and $g$ denotes the constant gray intensity applied to occluded pixels. Considering we use 8-bit RGB images, we set $g = 127$ for all channels, corresponding to a mid-level gray tone. This procedure removes distracting appearance artifacts introduced by occlusion, while preserving the visible, non-occluded regions of the object. The resulting masked images serve as input for downstream tasks, in our case being classification. As shown in Figure~\ref{fig:mask_comparison}, the effect of different threshold values $\tau$ on the masking is clearly visible.

\subsection{Occlusion severity estimation}
Beyond binary occlusion segmentation, we also quantify the severity of occlusion, defined as the proportion of the image affected by occlusion. We estimate the occlusion severity $\hat{s} \in [0,1]$ directly from the anomaly map $A$ by taking the mean of its pixel-wise anomaly scores:
\begin{equation}
    \hat s = \frac{1}{H \cdot W} \sum_{i=1}^{H} \sum_{j=1}^{W} A_{i,j}.
\end{equation}
The estimated severity $\hat s$ is subsequently used to select the most appropriate classification model for the given level of occlusion.

\subsection{Finetuning for occlusion robustness}
Empirically, we observed that models finetuned on a range of occlusion levels $[0, p]$ achieve their peak performance when evaluated on test images with occlusion severities close to $p$. 
At the same time, these models maintain competitive performance on images with lower occlusion severities ($< p$), indicating that exposure to a moderate range of occlusions promotes broader robustness.

To obtain models specialized for different levels of occlusion, we finetune multiple instances of the base model on synthetically occluded datasets. 
For each maximum occlusion level $p \in \{0, 10, \dots, 100\}$, we construct a corresponding training dataset $\mathcal{D}_{[0,p]}$ by applying gray occlusions with severities uniformly sampled from the range $[0, p]$. 
Each image in $\mathcal{D}_{[0,p]}$ is occluded with neutral gray regions covering a random fraction $p'$ of its area, where $p' \sim \mathcal{U}(0, p)$. 
Finetuning the base model on $\mathcal{D}_{[0,p]}$ yields a model denoted by $f_{[0,p]}$.

This approach exposes each model to a range of occlusion severities up to $p\%$, allowing it to adapt its feature representations accordingly. 
Furthermore, by applying a similar gray-masking at inference (guided by the estimated occlusion map) we simulate the visual conditions encountered during testing, thereby promoting consistency between training and inference.

\subsection{Severity-informed model selection}
We observed that models finetuned on datasets with specific occlusion ranges (e.g., $[0, p]$) tend to perform best on test images whose occlusion severity lies near the upper bound of that range. 
Performance gradually degrades as the test occlusion level deviates from the training range, suggesting that a single model may not perform optimally across the full spectrum of occlusion severities. 
We further investigate this behavior in the Experiments section.

To address this limitation, we maintain a pool of finetuned models
\[
\mathcal{F} = \{ f_{[0,p]} \mid p \in \mathcal{P} \},
\]
where $\mathcal{P}$ denotes the set of maximum occlusion levels used during finetuning. 
During inference, we estimate the occlusion severity of an input image as $\hat{s} \in [0,1]$.% (see Section~\ref{sec:occlusion-severity-estimation})

We then select the most suitable model $f_{[0,p^*]}$ from $\mathcal{F}$ based on the estimated severity $\hat{s}$:
\begin{equation}
    p^* = \arg\min_{p \in \mathcal{P}} |\hat{s} - p|,
\end{equation}
\begin{equation}
    f^* = f_{[0,p^*]}.
\end{equation}
This severity-informed selection ensures that each image is processed by the model best suited to its occlusion severity, thereby improving classification performance across varying levels of object visibility.

\section{Experimental Setup}\label{chap:experiments}

\subsection{Dataset}

We evaluate our method on an unoccluded finegrained dataset with class-level labels only, the Stanford Cars dataset~\cite{krause20133d}, which contains 196 car categories.
To study occlusion effects, we synthesize occluded images by applying Perlin noise \cite{perlin1985image} to generate smooth, spatially coherent masks that define the occluded pixel area and ratio. These masks are then filled with realistic cutouts of foliage, smoke, or rubble, extracted from natural images using the Segment Anything Model (SAM) \cite{kirillov2023segment}. We refer to these as textured occlusions, while non-textured occlusions are created by overlaying a uniform gray mask.

\subsection{Model architecture and training scheme}
For all experiments, we use the same finegrained classification model to ensure comparability. The model consists of a DINOv2 ViT-B/14 backbone and a multilayer perceptron (MLP) head. The 768-dimensional DINOv2 embeddings are passed through a hidden layer of 512 units with ReLU activation and dropout ($p=0.2$), followed by a linear output layer projecting to $C$ classes:
\begin{equation}
h(x) = \text{Linear}_{768 \rightarrow 512} \;\; \rightarrow \;\; \text{ReLU} \;\; \rightarrow \;\; \text{Dropout}(0.2) \;\; \rightarrow \;\; \text{Linear}_{512 \rightarrow C}.
\end{equation}
Finetuning proceeds in three stages to prevent catastrophic forgetting~\cite{french1999catastrophic}:
epochs 1–5 train only the MLP head;
epochs 6–15 jointly train the head and the last three DINOv2 layers;
epochs 16–20 finetune the entire network.
Training uses the Adam optimizer, with learning rates of $5\times10^{-3}$ for the MLP head and $5\times10^{-5}$ for the backbone.
The model is trained across varying occlusion severities.
In AnomalyDINO, the main hyperparameter is the memory bank size, i.e., the number of reference images in the patch feature bank $\mathcal{M}$. We populate $\mathcal{M}$ with non-occluded embeddings by selecting the training image nearest to each class centroid, as a single reference per class proved sufficient for stable performance.

\subsection{Evaluation metrics}
We want to know how well a model performs under occlusion. To quantify this, we measure the model's accuracy under increasing levels of occlusion and summarize it using the Area Under the Curve (AUC). This metric captures how well the model maintains performance as occlusion increases: higher values indicate better overall robustness. Let $p_0, p_1, \dots, p_n$ denote the discrete occlusion levels applied to the input images, and let $\mathrm{Acc}(p_i)$ be the classification accuracy at occlusion level $p_i$. The Area Under the Curve (AUC) of the accuracy-under-occlusion curve, $\text{AUC}_\text{occ}$. Furthermore, for the occlusion segmentation task, we use threshold-independent metrics that are widely used in binary segmentation. Specifically, we report the Area Under the Receiver Operating Characteristic (AUROC), as well as the Average Precision (AP).

\section{Findings}\label{chap:results}

\subsection{Occluders are distractors that deteriorate performance}

\begin{figure}[h]
    \centering
    \includegraphics[width=0.95\linewidth]{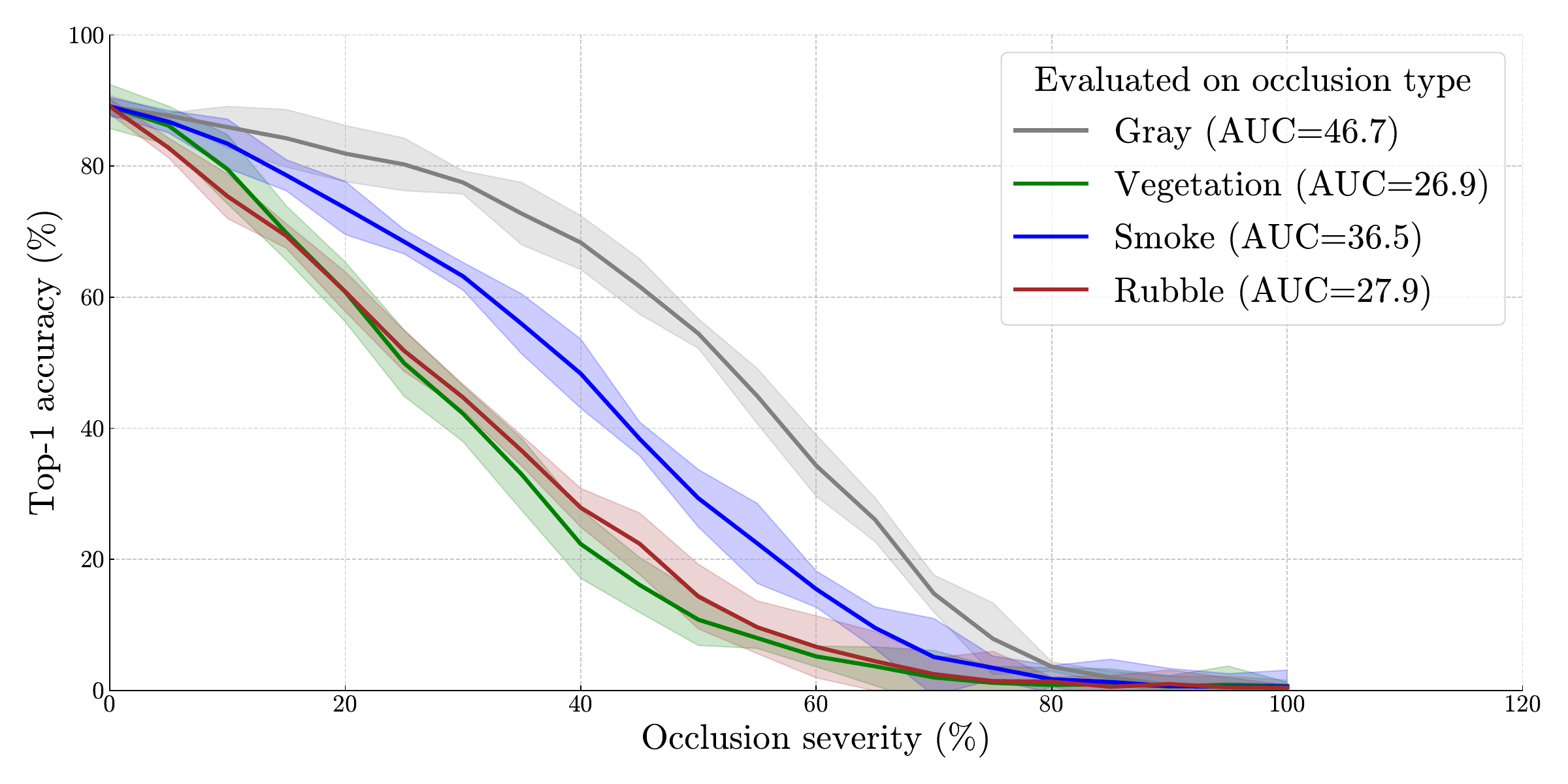}
    \caption{
    Larger degrees of occlusions (severity) deteriorate the performance of finegrained classification, trained without occlusions. Textured occlusions (vegetation, rubble) are more problematic than dull occlusion (smoke) or gray occlusion.
    }
    \label{fig:occl_type_comparison}
\end{figure}

A model that was trained without occlusions has problems with occlusions, as expected. We observe a large difference between gray-occluded versus texture-occluded images. Gray occlusions are simple uniform overlays, while textured occlusions involve more complex patterns such as vegetation or smoke. As shown in Figure~\ref{fig:occl_type_comparison}, textured occlusions disrupt the performance more severely than uniform gray ones. Vegetation and rubble are more problematic than smoke. This observation that occlusions distract is motivating our approach to finetune models specifically on gray-occluded data and applying gray masking at test time to enhance robustness.

\subsection{Occluders move the attention away from objects}

\begin{figure}[h]
    % \vspace{7.5mm}
    \centering
    \includegraphics[width=0.98\textwidth]{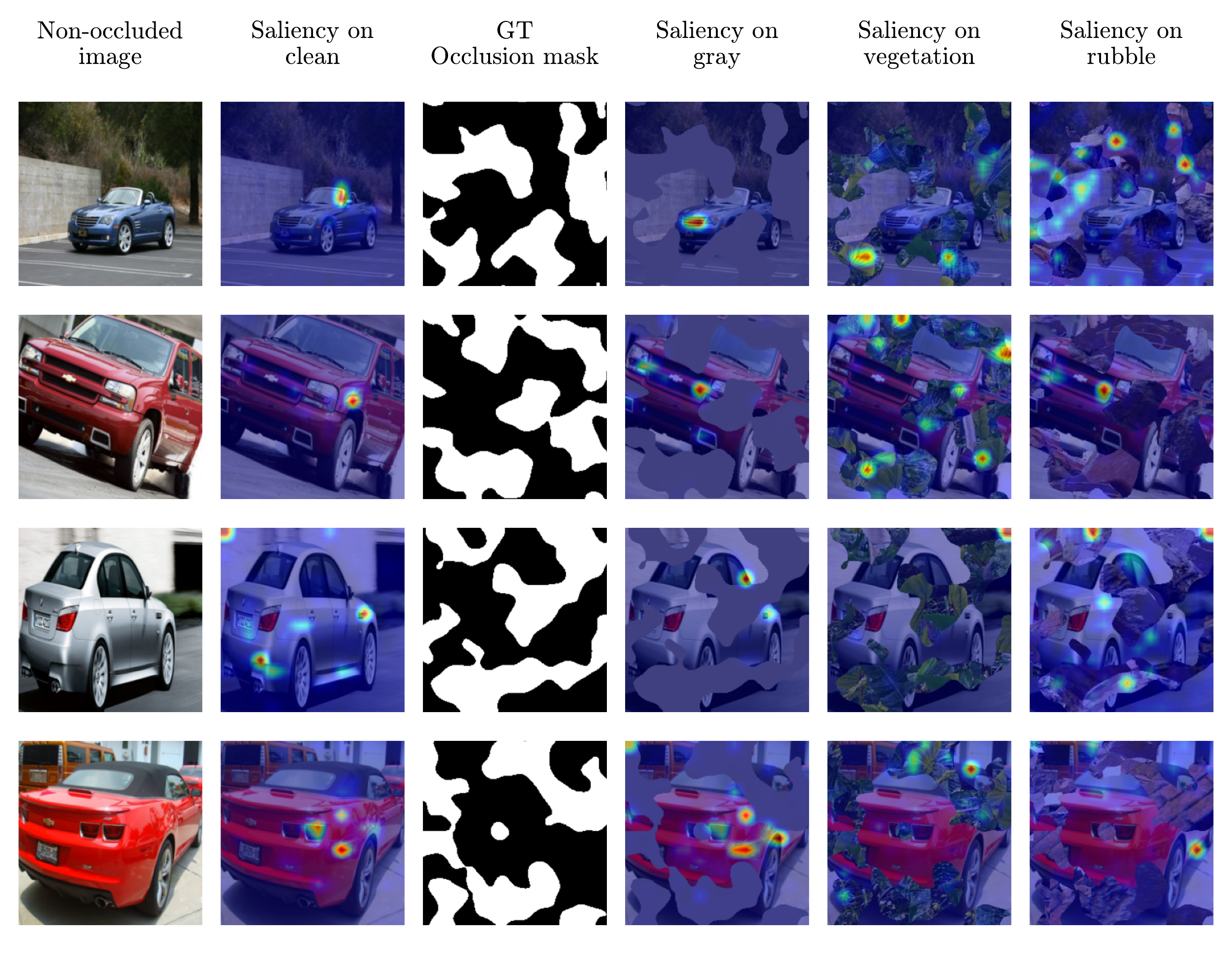}
    \caption{
    Textured occluders draw away the attention from the object. The first column shows the original (unoccluded) images, and the second column displays their corresponding saliency maps. The occlusion mask applied to each row is shown in column 3 and remains the same across all occlusion types. Columns 4–6 present the occluded images: gray, vegetation and rubble. Each is overlaid with its respective attention map. 
    }
    \label{fig:saliency_grid}
\end{figure}

To analyze the model’s attention under occlusion, we compare its focus across gray and textured occluded images. Using EigenGrad-CAM~\cite{muhammad2020eigen}\footnote{Implemented with the \texttt{pytorch-grad-cam} library~\cite{jacobgilpytorchcam}.} on the final layer of the DINOv2 feature extractor, we visualize high-level attention maps for both clean and occluded inputs, using the clean images as a reference baseline. As shown in Figure~\ref{fig:saliency_grid}, the model maintains relatively stable attention under gray occlusion, with saliency concentrated on the visible parts of the object. In contrast, textured occlusions (vegetation and rubble) disrupt this focus, often drawing attention toward the occluded regions themselves. These results suggest that textured occlusions interfere more strongly with the model’s ability to localize relevant object features compared to uniform gray occlusions.

\subsection{Occlusion can be localized and masked away}

\begin{table}[h]
    \centering
    \setlength{\tabcolsep}{6pt}
    \caption{
        Across various types of occlusions, the occlusion-agnostic AnomalyDINO performs better than OVSeg when prompted with the occlusion type. Mean AUROC (mAUROC) and mean Average Precision (mAP) is reported across occlusion severities (in rows) for occlusion types (vegetation, smoke, rubble; in columns).
    }
    \resizebox{1\linewidth}{!}{
    \begin{tabular}{lcccccc}
    \toprule
      
    \multirowcell{2}[-1ex][l]{\textbf{Method}}
    & \multicolumn{2}{c}{\textbf{Vegetation}} & \multicolumn{2}{c}{\textbf{Smoke}} & \multicolumn{2}{c}{\textbf{Rubble}}  \\
    \cmidrule(lr){2-3} \cmidrule(lr){4-5} \cmidrule(lr){6-7}
    & mAUROC & mAP & mAUROC & mAP & mAUROC & mAP \\
    \midrule
    & \multicolumn{6}{c}{\textbf{20\% Occluded}} \\
    \midrule
    AnomalyDINO & \textbf{95.64{\scriptsize$\pm$1.51}} & \textbf{79.35{\scriptsize$\pm$7.46}} & \textbf{94.46{\scriptsize$\pm$2.33}} & \textbf{76.59{\scriptsize$\pm$8.09}} & \textbf{94.40{\scriptsize$\pm$2.34}} & \textbf{74.56{\scriptsize$\pm$8.54}} \\
    OVSeg & \underline{80.38{\scriptsize$\pm$9.09}} & \underline{63.25{\scriptsize$\pm$15.26}} & 49.75{\scriptsize$\pm$5.55} & 24.16{\scriptsize$\pm$8.38} & 51.24{\scriptsize$\pm$6.30} & 25.54{\scriptsize$\pm$9.05} \\
    \midrule
    & \multicolumn{6}{c}{\textbf{40\% Occluded}} \\
    \midrule
    AnomalyDINO & \textbf{93.02{\scriptsize$\pm$3.33}} & \textbf{88.02{\scriptsize$\pm$5.76}} & \textbf{91.76{\scriptsize$\pm$3.51}} & \textbf{85.52{\scriptsize$\pm$6.24}} & \textbf{90.31{\scriptsize$\pm$4.47}} & \textbf{82.21{\scriptsize$\pm$7.63}} \\
    OVSeg & \underline{81.13{\scriptsize$\pm$11.35}} & \underline{75.28{\scriptsize$\pm$13.30}} & 52.26{\scriptsize$\pm$9.14} & 45.14{\scriptsize$\pm$8.55} & 51.18{\scriptsize$\pm$7.34} & 43.80{\scriptsize$\pm$6.66} \\
    \midrule
    & \multicolumn{6}{c}{\textbf{60\% Occluded}} \\
    \midrule
    AnomalyDINO & \textbf{86.83{\scriptsize$\pm$7.55}} & \textbf{89.03{\scriptsize$\pm$6.33}} & \textbf{87.80{\scriptsize$\pm$5.68}} & \textbf{89.33{\scriptsize$\pm$5.14}} & \textbf{82.34{\scriptsize$\pm$8.61}} & \textbf{83.95{\scriptsize$\pm$7.25}} \\
    OVSeg & \underline{78.54{\scriptsize$\pm$11.49}} & \underline{81.68{\scriptsize$\pm$9.05}} & 52.46{\scriptsize$\pm$10.10} & 63.15{\scriptsize$\pm$6.65} & 50.50{\scriptsize$\pm$6.07} & 61.25{\scriptsize$\pm$3.98} \\
    \midrule
    & \multicolumn{6}{c}{\textbf{80\% Occluded}} \\
    \midrule
    AnomalyDINO & 66.77{\scriptsize$\pm$16.14} & 88.43{\scriptsize$\pm$6.29} & \textbf{75.87{\scriptsize$\pm$13.15}} & \textbf{91.75{\scriptsize$\pm$4.88}} & \textbf{58.44{\scriptsize$\pm$17.11}} & \textbf{84.11{\scriptsize$\pm$6.87}} \\
    OVSeg & \underline{\textbf{73.93{\scriptsize$\pm$11.09}}} & \underline{\textbf{88.81{\scriptsize$\pm$4.78}}} & 52.75{\scriptsize$\pm$8.38} & 81.38{\scriptsize$\pm$3.43} & 51.85{\scriptsize$\pm$6.39} & 80.27{\scriptsize$\pm$2.29} \\
    \bottomrule
    \end{tabular}
    }
    \label{tab:ap_auroc_summary}
\end{table}

In order to perform gray masking and occlusion-severity estimation, it is essential to accurately localize the occluded regions within each image. We evaluate occlusion segmentation under three occluder types: vegetation, smoke, and rubble. Our method leverages AnomalyDINO for unsupervised occlusion segmentation, and we compare it against OVSeg, a prompt-based segmentation baseline that requires prior knowledge of the occlusion type (e.g., the text prompt “vegetation”). As summarized in Table~\ref{tab:ap_auroc_summary}, AnomalyDINO consistently outperforms OVSeg across all occlusion types. Unlike OVSeg, which performs well only for the occlusion it was prompted with, AnomalyDINO is occlusion-agnostic, accurately segmenting diverse occlusion types without any explicit supervision or prior knowledge about their nature.

\subsection{Occlusion-severity model selection is helpful} 

Having established that occlusions can be reliably segmented irrespective of their type, we evaluate how OASIC improves classification robustness under severe occlusions. Gray masking is applied using the occlusion maps obtained from AnomalyDINO, thresholded with Otsu’s method ($\tau_\text{Otsu}$), which we found to be near-optimal without requiring tuning. For estimating occlusion severity, we found that simply taking the mean of the pixel-wise occlusion likelihood map is sufficient. 

\begin{figure}[!ht] 
    \centering
    \includegraphics[width=\linewidth]{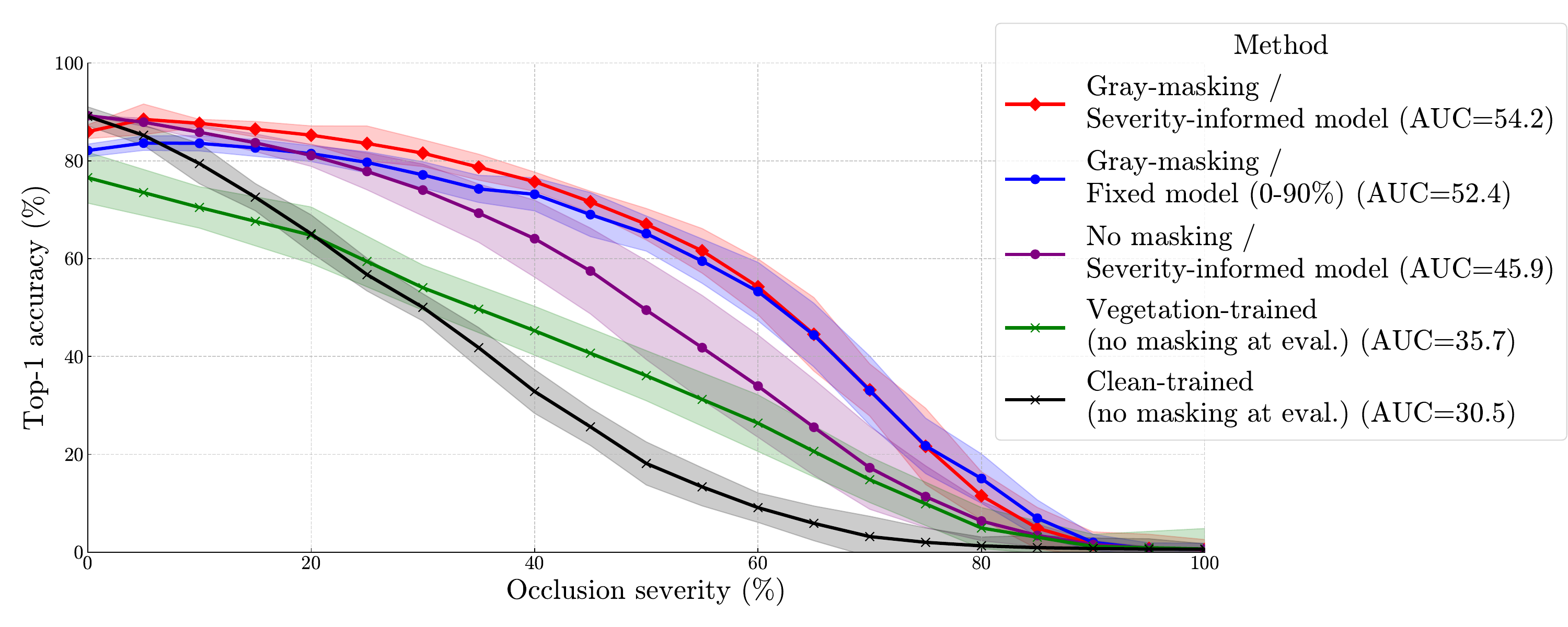}
    \caption{
        OASIC is much more robust to severe occlusions, improving 5x compared to standard training. Both the gray-masking of occlusion (to suppress distraction) and the severity-informed model selection (to handle limited visibility) contribute to OASIC's performance.
    }
    \label{fig:occlusion_robustness_comparison}
\end{figure}

We evaluate five configurations in Figure~\ref{fig:occlusion_robustness_comparison} to isolate the contributions of suppressing the occluder's distraction vs. severity-informed model selection, and their combination. \textcolor{red}{Red} represents our full method (OASIC), combining gray masking with severity-informed model selection. \textcolor{blue}{Blue} applies only gray masking using a fixed model trained across a wide occlusion range (0–90\%), isolating the masking effect. \textcolor{purple}{Purple} uses only severity-informed model selection without masking, assessing the benefit of adaptive model choice. \textcolor{green}{Green} denotes a model trained on vegetation occlusion without masking, serving as an occlusion-specific baseline. Finally, \textcolor{black}{black} corresponds to a model trained on unoccluded images, representing the naive baseline with no occlusion handling.

OASIC (\textcolor{red}{red}) structurally achieves the highest robustness and overall performance across all occlusion severities. By integrating occlusion localization, gray masking, and severity-informed model selection, it effectively mitigates the impact of occlusion. In terms of $\text{AUC}_\text{occ}$, OASIC outperforms the vegetation-train configuration (\textcolor{green}{green}) by $+18.5$ and the clean-train configuration (\textcolor{black}{black}) by $+23.7$, attaining the highest $\text{AUC}_\text{occ}$ among all evaluated methods.

\section{Conclusion}\label{chap:discussion}

We investigated finegrained image classification under severe occlusions. Our analyses showed that textured occluders degraded accuracy far more than uniform gray ones, highlighting that the visual properties of occluders could mislead models. We found that no single model performed optimally across varying levels of occlusion severity. We introduced OASIC, which leveraged pixel-wise occlusion likelihoods to detect and neutralize occluded regions by replacing them with a uniform gray mask. This approach suppressed distracting visual cues while preserving relevant object information. In addition, we finetuned a pool of classification models and dynamically selected the most appropriate model based on the estimated occlusion severity. By combining gray masking with severity-informed model selection, OASIC provided a robust and adaptive solution for diverse occlusion conditions. Quantitatively, our method improved $\text{AUC}_\text{occ}$ by $+18.5$ compared to training on occluded images directly and by $+23.7$ compared to fine-tuning on unoccluded images, demonstrating a substantial gain in occlusion robustness.

The proposed framework has potential beyond finegrained recognition. Tasks involving partial visibility, such as autonomous navigation, visual surveillance, or medical imaging, could similarly benefit from anomaly-based occlusion estimation. Extending the approach to these domains would further demonstrate its generality and robustness. Future work could integrate occlusion estimation and classification within a single architecture, using the pixel-wise occlusion likelihoods as an additional supervision signal. This would enable models to learn to reason about occlusion directly and adapt to varying levels of visibility in a unified, end-to-end manner.

\bibliographystyle{splncs04}
\bibliography{references}

\end{document}